\documentclass[pre,aps,twocolumn,showpacs,amsmath,amssymb]{revtex4}
\usepackage{graphicx}

\usepackage{subfigure}

\newtheorem{theorem}{Theorem}

\begin{document}

\title{Solving the undirected feedback vertex set problem by local search}

\author{Shao-Meng Qin and Hai-Jun Zhou\footnote{Corresponding author.
Email address: {\tt zhouhj@itp.ac.cn}.}}

\affiliation{State Key Laboratory of Theoretical Physics, 
Institute of Theoretical Physics, Chinese Academy of Sciences, Beijing 
100190, China}

\date{\today}

\begin{abstract}
  An undirected graph consists of a set of vertices and a set of undirected
  edges between vertices.  Such a graph may contain an abundant number of
  cycles, then a feedback vertex set (FVS) is a set of
  vertices intersecting with each of these cycles.
  Constructing a FVS of cardinality approaching the global minimum value is a
  optimization problem in the nondeterministic 
  polynomial-complete complexity class, therefore it might be
  extremely difficult for some large graph instances. 
  In this paper we develop a simulated annealing local search algorithm for the
  undirected FVS problem.
  By defining an order for the vertices outside the FVS, we replace the global
  cycle constraints by a set of local vertex constraints on this order. Under
  these local constraints the 
  cardinality of the focal FVS is then gradually reduced by the simulated
  annealing dynamical process.
  We test  this heuristic algorithm on large instances of Er\"odos-Renyi random
  graph and regular random graph, and find that this algorithm is comparable in
  performance to the  belief propagation-guided decimation algorithm.
\end{abstract}

\pacs{89.20.Ff, 02.70.Uu, 02.10.Ox, 75.10.Nr}

\maketitle

%
%

\section{Introduction}

An undirected graph is formed by a set of vertices and a set
of undirected edges, with each edge connecting between two different vertices.
A feedback vertex set (FVS) for such a graph is a set of vertices
intersecting with every cycle of the graph. In other words, the subgraph
induced by the vertices outside the FVS contains no cycle (it is a forest) 
\cite{Garey-Johnson-1979,Festa-Pardalos-Resende-1999}. 
The feedback vertex set problem aims at constructing a FVS
of small cardinality for a given undirected graph.
It is a fundamental nondeterministic polynomial-complete
(NP-complete) combinatorial optimization problem with global cycle constraints
\cite{Garey-Johnson-1979,Karp-1972,Cook-1971}.
In terms of complete algorithms, whether a graph $G$ has a FVS of cardinality
smaller than $n$ can be determined in time 
$O(3.592^{n})$ \cite{Cao-Chen-Liu-2010,Kociumaka-Pilipczuk-2013,Guo-etal-2006}.
And an FVS of cardinality at most two times the optimal value can be
easily constructed by an efficient polynomial algorithm
\cite{Bafna-Berman-Fujito-1999}.
An optimal FVS is a feedback vertex set whose cardinality is
the global minimum value among all the feedback vertex sets of the graph.
For a given graph, an optimal FVS can be constructed in an exact way in 
time  $O(1.7548^{N})$ \cite{Razgon-2006,Fomin-Gaspers-Pyatkin-2006}, 
where $N$ denotes the total number of vertices in the graph.
Applied mathematicians have obtained rigorous lower and upper bounds for the
optimal FVS problem \cite{Bau-Wormald-Zhou-2002} and have proved its
tractability for graphs with specific structures (see for example, 
\cite{Wang-Wang-Chang-2004,Jiang-Liu-Xu-2011}  and references cited therein).

Due to the NP-complete nature of the FVS problem, 
in general it is not feasible to construct optimal feedback vertex sets
for large cycle-rich graphs. An important question is then to
design efficent heurstic algorithms that are able to obtain near-optimal
FVS solutions for given graph instances. Such a task is quite nontrivial.
A major technical difficulty is that cycles are global objects of a
graph and therefore the existence of cycles can not be judged by
checking only the neighborhood of a vertext. (Similar difficulties exist
in other combinatorial optimization problems with global constraints,
such as the Steiner tree problem \cite{Bayati-etal-2008} and the optimal
routing problem \cite{Yeung-Saad-Wong-2013,Yeung-Saad-2013b}.)
In Ref.~\cite{Zhou-2013}, one of the authors succeeded in converting the FVS
problem to a spin glass problem with local interactions. The FVS problem
was then studied from the spin glass perspective, and
a message-passing algorithm, belief propagation-guided decimaton (BPD), 
was impletmented to solve the FVS problem heuristically for single graph
instances. This BPD algorithm is quite efficient in terms of computing time and
computer memory (since there is no need of cycle checking),
and it can obtain FVS solutions that are very close to the
optimal ones when applied on large random graph instances and
regular lattices \cite{Zhou-2013}.

For the undirected FVS problem it is not yet known whether simple local 
search algorithms can achieve equally excellent results  as the BPD algorithm.
Motivated by this question, we complement the
message-passing approach  in this paper by 
implementing and testing a simulated annealing local
searching (SALS) protocol 
\cite{Kirkpatrick-etal-1983} for the undirected FVS problem.
A similar algorithmic study has already been undertaken in
\cite{Galinier-Lemamou-Bouzidi-2013} for directed graphs.
 Here we  modify the microscopic search rules of 
\cite{Galinier-Lemamou-Bouzidi-2013} to make it applicable to undirected
graphs.
In the SALS algorithm, an order is defined for the vertices outside the
focal FVS, and this order is constrained by a set of local vertex constraints.
Our simulation results suggest that
this local search algorithm is comparable in performance to the BPD
algorithm at least for random graph instances.

The feedback vertex set problem has wide practical applications in
the field of computer science (such as integrated circuit design and database
management).
Although not yet seriously explored, the FVS problem may have many potential
applications in complex systems research as well. 
For example, if a vertex is contained in a large  fraction of the near-optimal
feedback vertex sets, we may expect this vertex to play a very significant role 
for the dynamical processes on the graph. Therefore the probability of
belonging to a near-optimal FVS can serve as a centrality index of dynamical
significance for each vertex of
a graph. Such a probablity can be computed by sampling many independent
near-optimal FVS solutions or be computed directly using the belief propagation
iterative equations \cite{Zhou-2013}.

\begin{figure}
  \begin{center}
    \includegraphics[width=0.35\textwidth]{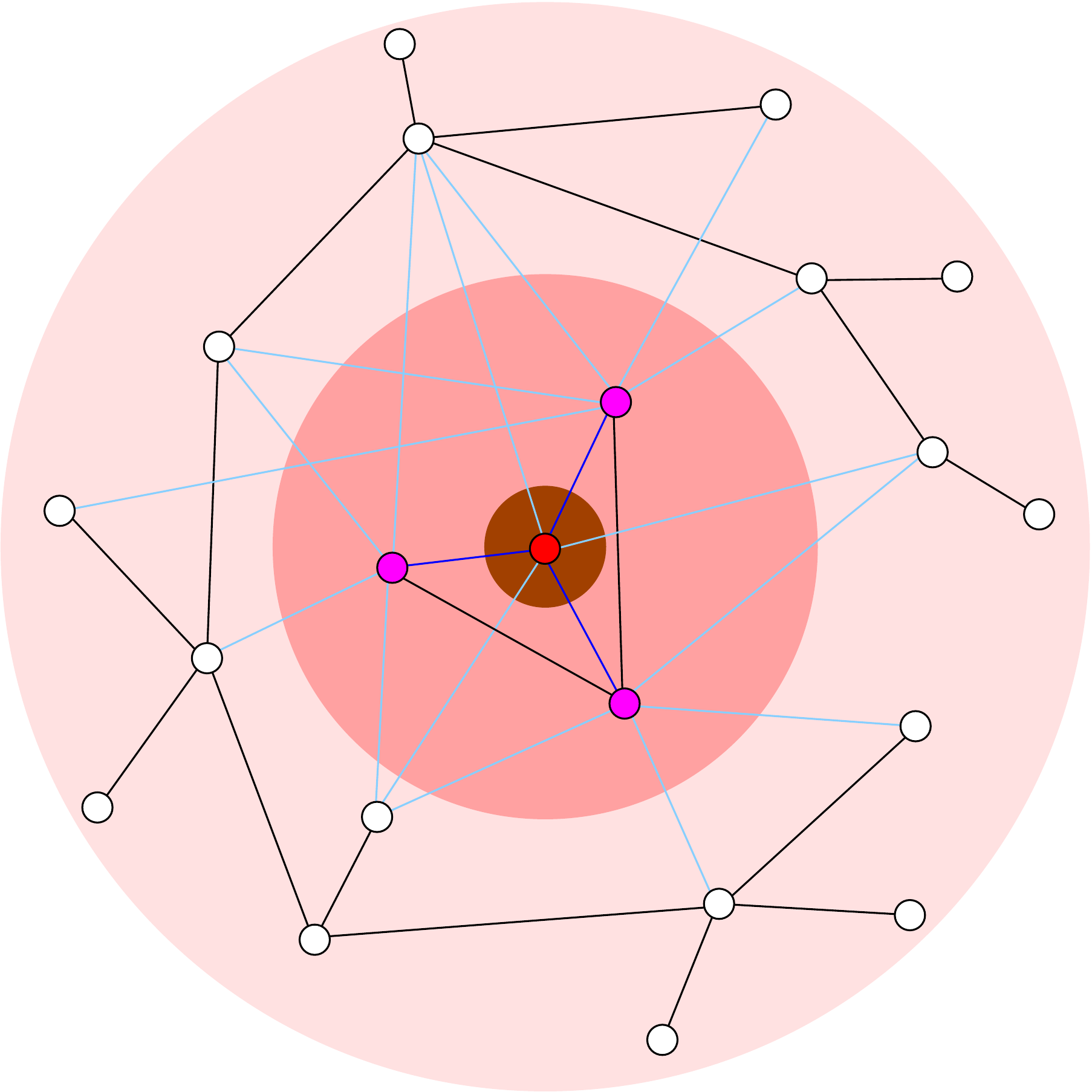}
  \end{center}
  \caption{
    \label{fig:FVShierarchy}
    (color online). The four  filled  points form an
    optimal feedback vertex set for this graph $G$. The subgraph $g$
    induced by these four vertices and the cycle-free subgraph induced by all the
    remaining vertices (shown as open points) are connected through many
    edges (shown in light blue). Since the subgraph $g$ still contains cycles
    within itself, we decompose it into a tree subgraph of three
    vertices (filled magenta points) and a subgraph formed by a FVS of
    one vertex (the central red point).
    By this way, the vertices in the orginal graph $G$ are arranged into
    three different layers. The vertices of each layer form a cycle-free
    subgraph (a tree or a forest), while different layers are connected by
    edges. An important property of such an organization is that each cycle
    must involves vertices from at least two layers.
  }
\end{figure}

The construction of a near-optimal FVS also facilitates the study of
a complex dynamical system as a simpler response
problem of a cycle-free subsystem (which is
intrinsically simple) under the influence of the vertices in  the FVS vertices.
If the subgraph induced by the vertices in the FVS itself contains many
cycles, such a decomposition can be applied on this subgraph again. Through this
iterated process, an initial cycle-rich complex graph is then organized 
into a hierarchy of forest (cycle-free) subgraphs and the edges between these
forests. A simple illustration of this hierarchical organization is shown in
Fig.~\ref{fig:FVShierarchy}.
We believe such a hierarchical representation of a complex graph
will be very helpful in future dynamical applications.

The next section describes the SALS algorithm in detail, and in
Sec.~3 we test the performance of this local search algorithm
on random graph instances and compare the results with the results obtained
by the BPD algorithm and those obtained by the replica-symmetric mean
field theory. We conclude this work in Sec.~4. The two appendices are
the proofs of the theorems of Sec.~2.

%
%
\section{The local search algorithm}

For a graph $G$ of $N$ vertices, let us consider an ordered list
$L$ formed by $n \leq N$ vertices of this graph
\begin{equation}
\label{eq:gList}
L \equiv (v_1, v_2, \ldots, v_n) \; .
\end{equation}
Following \cite{Galinier-Lemamou-Bouzidi-2013},
we assign to the first vertex $v_1$ of this list an
integer rank $r=1$, to the second vertex an integer rank $r=2$,
..., and to the last vertex $v_n$ an integer rank $r=n$. Therefore
each vertex $i\in L$ has an integer rank $r_i$ which marks the 
position of this vertex in the list $L$. For the purpose of the
FVS problem, we introduce for each
vertex $i\in L$ a ranking condition as
\begin{equation}
\label{eq:rc}
\sum\limits_{j: j\in L,  (i,j) \in G}
 \Theta(r_i-r_j) \leq 1 \; , 
\end{equation}
where $(i,j)$ denotes an edge of the graph $G$ between vertex $i$ and
vertex $j$, and $\Theta(x)=1$ if $x>0$ and $\Theta(x)=0$ if $x\leq 0$.
The ranking condition (\ref{eq:rc}) is satisfied by vertex $i$ if and
only if among all the nearest neighboring vertices of vertex $i$
that are also contained in the list $L$, at most one of them has a
lower rank than that of $i$.
A list $L$ is referred to as a legal list if all its vertices satisfy
the ranking condition (\ref{eq:rc}). The link between legal lists and
feedback vertex sets is setup by the following two theorems:
\begin{theorem}
\label{th:1}
If  $L$ is a legal list, then the subgraph of the
graph $G$ induced by all the vertices
of this list is cycle-free. Therefore the set $\Gamma$ formed by
all the  remaining vertices of $G$ not included in $L$ is a FVS.
\end{theorem}
\begin{theorem}
\label{th:2}
If $\Gamma$ is a FVS for a graph $G$, then it is possible to form
a legal list $L$ using all the vertices not contained in $\Gamma$.
\end{theorem}
These two theoretms are easy to prove, see the appendices for technical
details. They suggest that there is a one-to-many correspondance
between a FVS $\Gamma$ and legal lists $L$.
Therefore the problem of constructing an optimal FVS is converted to
a problem of constructing a legal list $L$ of maximal cardinality. 
Notice that judging whether a list $L$ is legal or not is algorithmically
very easy as Eq.~(\ref{eq:rc}) involves only the neighborhood a focal
vertex $i$ but not the connection pattern of the whole graph $G$. 
A similar conversion from global constraints to local constraints has
also been used in the Steiner tree problem, which aims at constructing
a tree of minimal total edge length connecting a set of specified vertices
\cite{Bayati-etal-2008,Biazzo-Braunstein-Zecchina-2012}.

Just for simplicity of later discussions, let us define the energy of a
legal list $L$ as the total number of vertices not contained in it,
namely
\begin{equation}
  \label{eq:Energy}
  E(L) \equiv N-  \sum\limits_{i\in L} 1 \; .
\end{equation}
In otherwords, $E(L)$ is just the cardinality of the complementary
FVS $\Gamma$ of the list $L$.
Following Ref.~\cite{Galinier-Lemamou-Bouzidi-2013} we implement a 
simulated annealing local search algorithm as follows:
\begin{enumerate}
\item[0.]
  Input the graph $G$. Initialize the legal set $L$ as containing
  only a single randomly chosen vertex of $G$, and the complementary
  feedback vertex set $\Gamma$ then contains all the other vertices of $G$. 
  Set the temperature $T$ to
  an initial value $T_0$.

\item[1.] 
  Choose a vertex (say $i$) uniformly at random from the feedback
  vertex set $\Gamma$.

  (a)
  If the list $L$ contains no nearest neighbor of the vertex $i$,
  delete $i$ from  $\Gamma$ and insert it to the head of $L$. 
  Then vertex $i$ has rank $r_i=1$ in the updated list $L$ and the energy
  of $L$ decreases by $1$.
  
  (b)
  If $L$ contains exactly one nearest
  neighbor (say $j$ with rank $r_j$) of the vertex $i$, 
  delete $i$ from $\Gamma$ and
  insert it to $L$ at the position just after vertex $j$. 
  Then vertex $i$ has rank $r_i=r_j+1$ in the
  updated list $L$ and the energy of $L$ decreases by $1$.
  
  (c)
  If $L$ contains two or more nearest neighbors (say $j, k, \ldots$ with
  vertex $j$ having the lowest rank $r_j$ among these
  vertices) of the
  vertex $i$, we make a proposal of moving $i$ from 
  $\Gamma$ to the ordered list $L$
  at the position just after vertex $j$ and deleting all 
  those nearest neighbors of $i$ from $L$ if the insertion of
  $i$ causes the violation of the
  ranking condition (\ref{eq:rc}) for these vertices. Suppose
  $n_d \geq 0$ vertices have to be deleted from $L$ (and
  be added to $\Gamma$) as a result of
  inserting vertex $i$ to $L$, then the energy increase of $L$ is
  $\Delta E = -1 + n_d$.
  If $\Delta E \leq 0$ we accept the proposed action  with
  probability $1$, otherwise we accept it  with probability
  $\exp(-\Delta E /T)$. If this proposal is accepted, then
  vertex $i$ has rank $r_i=r_j+1$ in the updated list $L$.
  
\item[2.] 
  Repeat step 1  until the list $L$ has been successfully updated
  for $N_t$ times. During this process, record the best FVS so far reached
  and the corresponding lowest energy, $E_{min}$.

\item[3.] 
  Decrease the temperature to  $T \leftarrow T \times \alpha$, where
  $\alpha<1$ is a fixed constant, and then repeat steps 1-2. 
  If the energy value $E_{min}$ does not change in  $N_{fail}$ contiguous
  temperature stages, then we stop the local search process and
  output the reached best FVS.
\end{enumerate}

In our computer experiments we use $T_0=0.6$, $N_t=50$ and $N_{fail}=50$ which
are identical to the values used in 
Ref.~\cite{Galinier-Lemamou-Bouzidi-2013}. The temperature
ratio $\alpha$ is set to several different 
values $\alpha=0.99$, $\alpha=0.98$ and
$\alpha=0.9$. As shown in Fig.~2, the slower the rate of temperature
decrease, the lower is the energy of the constructed best FVS solutions.

\section{Numerical results}

The BPD algorithm was applied to Erd\"os-Renyi (ER) random graphs and
regular (RR) random graphs in \cite{Zhou-2013} to test its performance.
As we want to  compare the performance of the present SALS algorithm with
the BPD algorithm, we apply the SALS algorithm on the same ER and
RR random graphs used in \cite{Zhou-2013}.

An ER random graph of $N$ vertices and $M = (c/2) N$ edges was generated
in \cite{Zhou-2013} by first selecting $M$ different vertex pairs
uniformly at random from the whole set of $N(N-1)/2$ candidate vertex pairs
and then connecting the two vertices of each selected vertex pair by
an edge. A vertex in such a graph has on average $c$ nearest neighbors (i.e.,
the mean vertex degree of the graph is $c$).
When $N \gg 1$ the vertex degree distribution of the graph converges to
a Poisson distribution of mean value $c$ 
\cite{He-Liu-Wang-2009}.
A RR graph differs from an ER graph with the additional constraint that
each vertex has exactly the same number $c$ of nearest neighbors (here
$c$ must be an integer). It was generated by first attaching to each
vertex a number of $c$ half-edges and then randomly connecting two half-edges 
into a full edge, but prohibiting self-loops or multiple edges between the same
pair of vertices (see Ref.~\cite{Zhou-Lipowsky-2007,Zhou-Lipowsky-2005}
for more details of the graph generation process).

The vertex number of all these random graph instances is equal to the same
value $N=10^5$. There are $96$ independently generated ER or RR random graph
instances at each fixed (mean) degree value $c$. We run the SALS process once
on each of these $96$ instances and then compare the mean value of the
obtained final $E_{min}$ values with the mean value of the final energies
obtained in \cite{Zhou-2013} by running the BPD process once on each of 
these $96$ graph instances.

\begin{figure}
  \begin{center}
    \subfigure[]{
      \label{fig:evolution:a}
      \includegraphics[width=0.35\textwidth,angle=270]{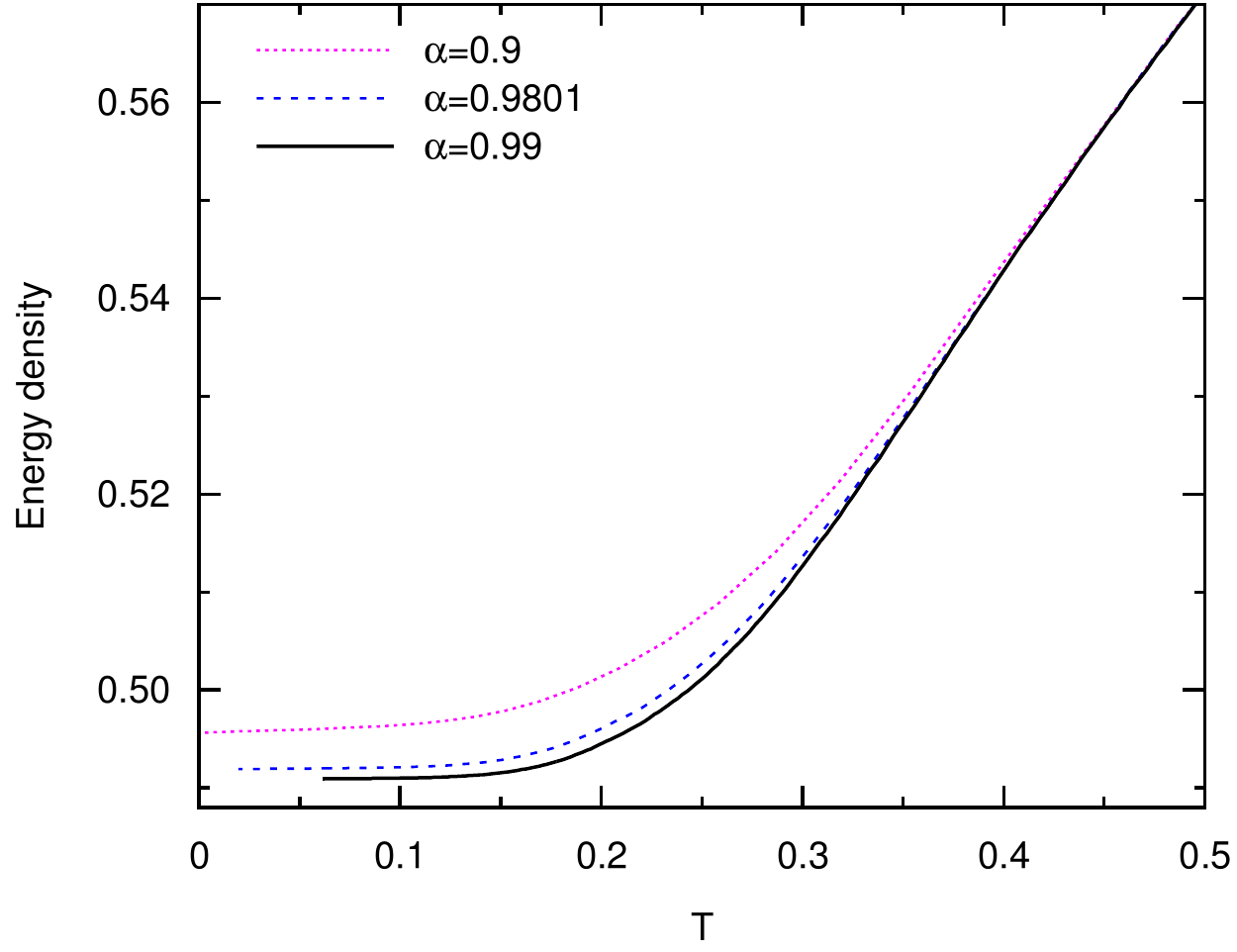}
    }
    \subfigure[]{
      \label{fig:evolution:b}
      \includegraphics[width=0.35\textwidth,angle=270]{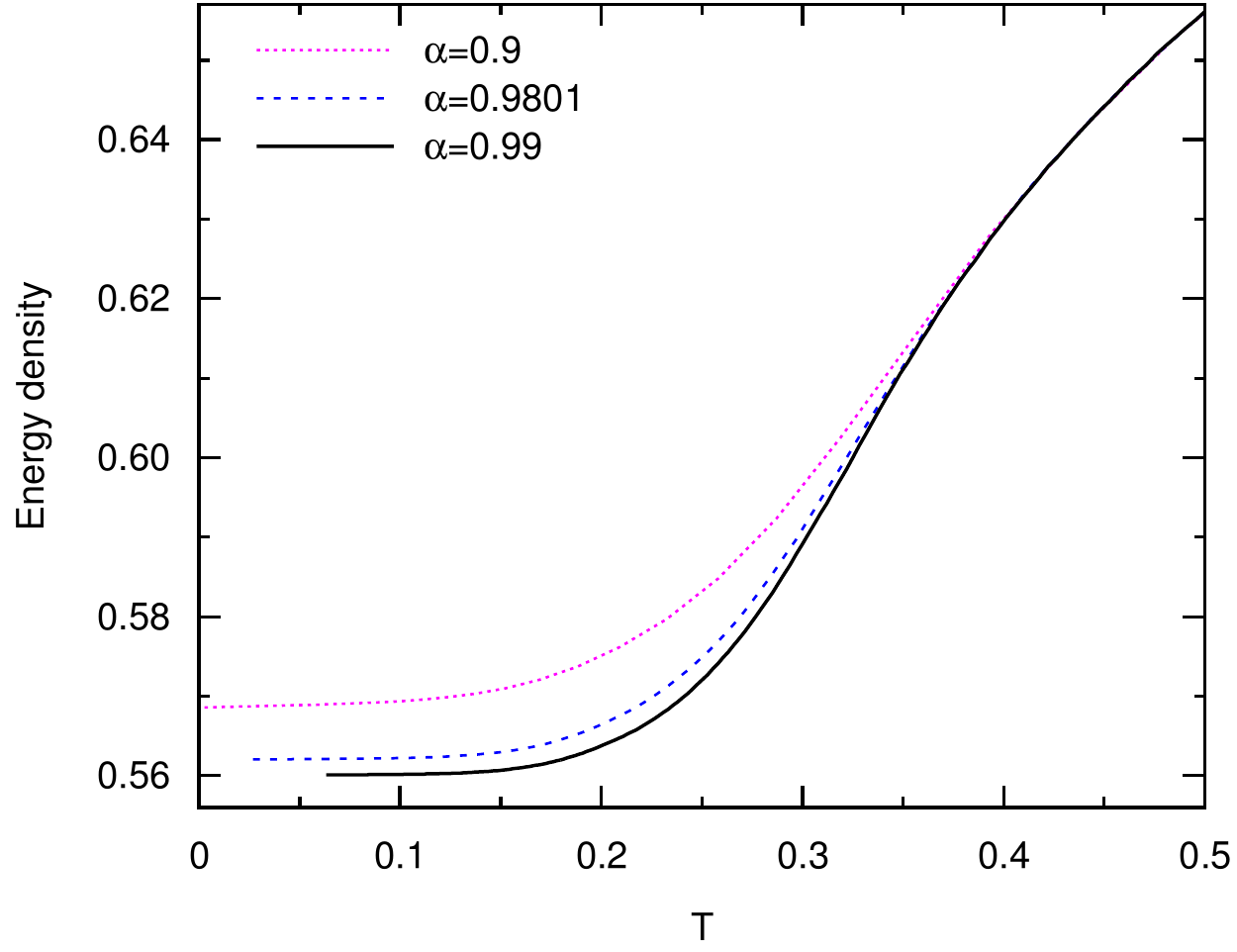}
    }
  \end{center}
  \caption{
    \label{fig:evolution}
    Evolution of the minimal energy density. The minimal energy density
    is equal to $E_{min}/N$, where $E_{min}$ is the lowest energy so far
    reached during the annealing process from the initial temperature
    $T_0=0.6$ to the current value $T$. Each curve is the averaged result
    of  a single run of the SALS algorithm on $96$ random graph instances
    of vertex number $N=10^5$.
    (a) ER random graphs of mean vertex degree $c=10$. (b) RR random 
    graphs of vertex degree $c=10$.
  }
\end{figure}

The evolution of the minimal energy $E_{min}$ as a function of the
temperature $T$ is shown in Fig.~\ref{fig:evolution} for ER (a) and
RR (b) random graphs of $c=10$.
When $T>0.4$ the evolution curves for the three different
cooling parameters $\alpha=0.9$, $0.9801$ and $0.99$ coincide with each other.
This indicates that at $T>0.4$ the typical relaxation time of energy is shorter
than the time scale of temperature decreasing. At $T \approx 0.4$ the curve of
$\alpha=0.9$ starts to separate from the other two curves, and the
final value of $E_{min}$ is considerably higher than those of the other
two evolutionary trajectories. This indicates that at cooling parameter
$\alpha=0.9$ the simulated annealing process is eventually trapped in a local
region of the configuration space whose energy minimal value is extensively
higher than that of the optimal solutions.
The two evolutionary curves of $\alpha=0.9801$ and $\alpha=0.99$ again
separate from each other at a lower temperature of $T \approx 0.35$, and
the final value of $E_{min}$ reached at cooling parameter $\alpha=0.99$ is
slightly lower than the final value of $E_{min}$ reached at $\alpha=0.9801$.
If the value of $\alpha$ is set to be more closer to $1$ the final value of
$E_{min}$ will decrease slightly further (at the expense of much longer
simulation times). The observation that lower final energy values can be
reached by lowering the cooling rate strongly indicates the system has a
very complicated low-energy landscape with many local minima.
This is consistent with the prediction that the
undirected FVS problem is in a spin glass phase at low enough
energy values \cite{Zhou-2013}.

\begin{figure}
  \begin{center}
    \subfigure[]{
      \label{fig:FVS:a}
      \includegraphics[width=0.35\textwidth,angle=270]{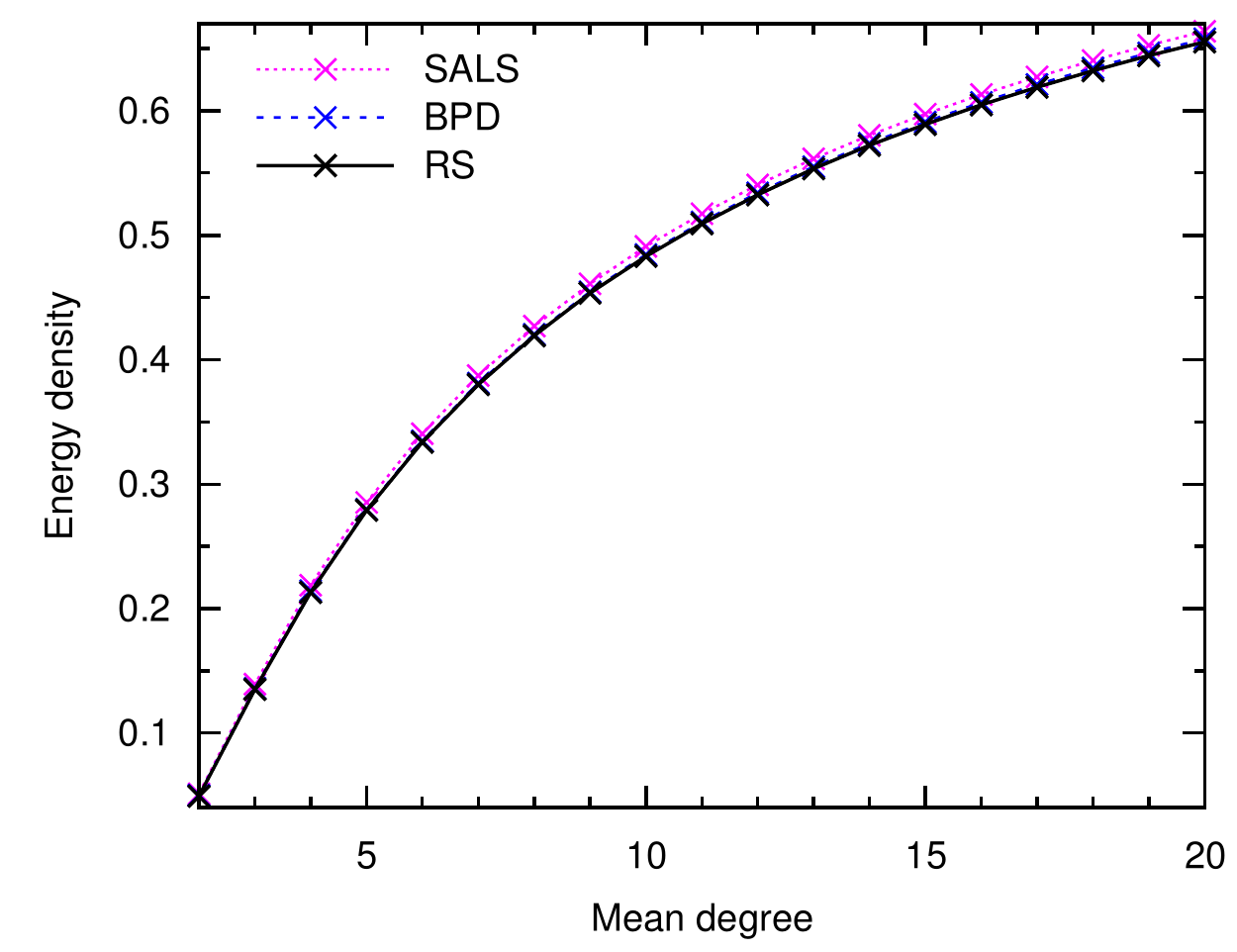}
    }
    \subfigure[]{
      \label{fig:FVS:b}
      \includegraphics[width=0.35\textwidth,angle=270]{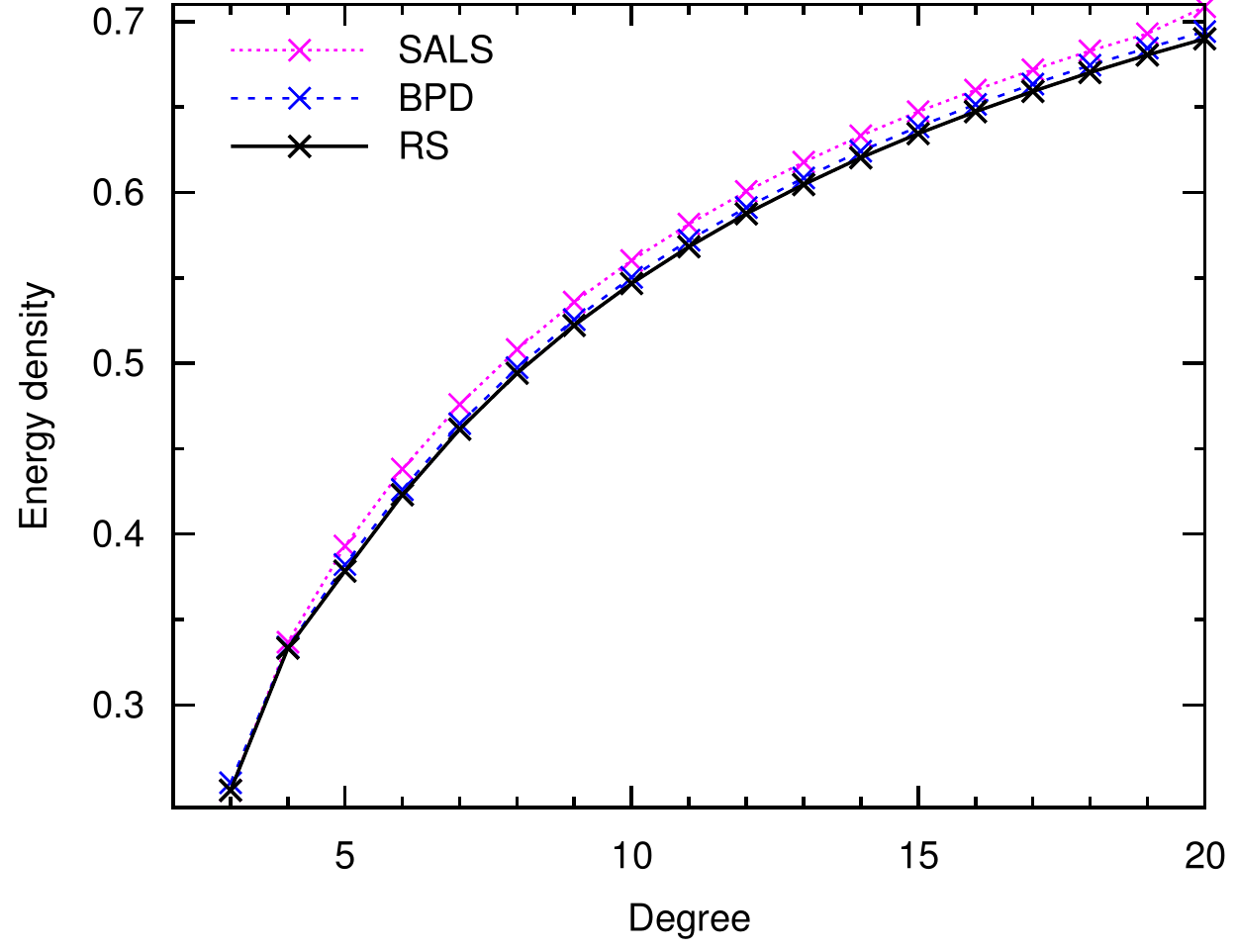}
    }
  \end{center}
  \caption{
    \label{fig:FVS}
    The energy density (i.e., relative size) of feedback vertex sets for
    ER (a) and RR (b) random graphs. 
    The SALS and BPD algorithms are applied to $96$ random graph instances of
    $N=10^5$ at each value of (mean) vertex degree $c$. The averaged results
    over these $96$ instances are shown here (the standard deviations are
    smaller than the size of the symbols and are not shown for clarity).
    As a comparison we also show the results obtained by the 
    replica-symmetric (RS) mean field theory at $N=\infty$.
  }
\end{figure}

Figure~\ref{fig:FVS} compares the performance of the SALS algorithm (the
cooling parameter fixed to $\alpha=0.99$) with that
of the BPD algorithm on ER (a) and RR (b) random graphs.
We see that BPD slightly outperforms SALS for both ensembles of random graphs.
This is not surprising, since the BPD algorithm takes into account the global
structure of the graph through message-passing, while the SALS algorithm
considers only the local graph structure. It is interesting to see that
the results of the two algorithms are actually very close to each
other, especially for ER random graphs.
As also shown in Fig.~\ref{fig:FVS}, at each value of (mean) degree $c$,
the result obtained by the SALS algorithm is very close to the predicted
value of the global minimal energy by the replica-symmetric mean field
theory \cite{Zhou-2013}. 

Therefore message-passing algorithms are not the only candidate choices to
construct near-optimal feedback vertex sets for undirected random graphs.
An advantage of the SALS algorithm is that its implementation is very 
straightforward. It may be the method of choice for many practical
applications.

\begin{figure}
  \begin{center}
    \includegraphics[width=0.35\textwidth,angle=270]{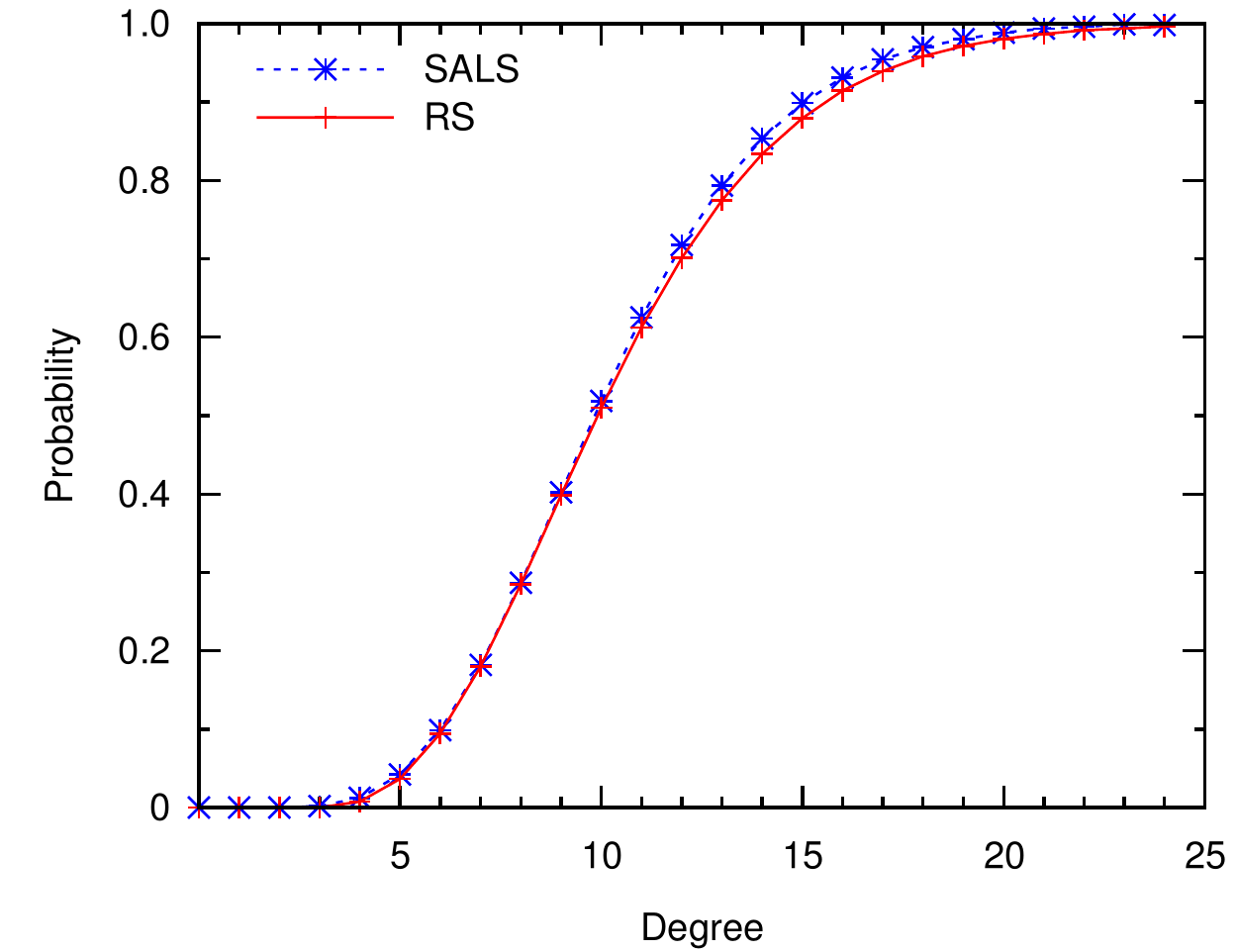} 
  \end{center}
  \caption{
    In an ER random graph of mean degree $c=10$,
    a vertex's probability of being in a feedback vertex set increases
    with its vertex degree $k$. Each star point is the mean result obtained
    by averaging over the $96$ final FVS solutions obtained by the SALS
    algorithm on the $96$ random graph instances of $N=10^5$ vertices. 
    Each plus point denotes
    the result obtained by the RS mean field theory at $N=\infty$
    \cite{Zhou-2013}.
  }
  \label{fig:vertexFVSprob}
\end{figure}

We also notice from Fig.~\ref{fig:FVS} that the SALS algorithm performs
poorer in regular random graphs than in ER random graphs. This is probably
due to the following fact: Each vertex in a 
regular random graph has the same degree so there is local guide
as to whether a vertex should be included into the
feedback vertex set or not. On the other hand the degree heterogeneity of
an ER graph will give some local guide to the SALS process to arrive
at a near-optimal FVS. Based on the final $96$ FVS solutions obtained by
the SALS algorithm on the $96$ ER random graphs with mean vertex 
degree $c=10$, we compute the mean probability $f_k$ that a vertex of
degree $k$ is contained in the FVS. The results shown in 
Fig.~\ref{fig:vertexFVSprob} demonstrate that $f_k$ is close to $0$ for
$k\leq 4$ and close to $1$ for $k\geq 20$, and it is
an rapidly increasing function of $k$ in the range of $5 \leq k \leq 15$.
The empirically obtained mean value of $f_k$ is found to be very close to the
value of $f_k$ computed using the replica-symmetric
mean field theory. Notice that for $10 \leq k \leq 20$
the empirical values of $f_k$ are slightly larger than those predicted by
the RS mean field. Such small differences cause the energies of the
construsted FVS solutions to be higher than the corresponding
optimal values.

The graphs for real-world complex systems usually are highly heterogenous
in terms of vertex degree distributions \cite{He-Liu-Wang-2009}. Therefore
it is very likely that the SALS algorithm to have very good performance for
such graphs.

\section{Conclusion}

In this paper, we implemented and tested a local search algorithm for the 
undirected feedback vertex set problem. Similar to the local search algorithm 
for the directed FVS problem \cite{Galinier-Lemamou-Bouzidi-2013}, our
algorithm uses the technique of simulated annealing to explore the 
highly complex landscape of low-energy congiurations. Our simulation
results demonstrated that this algorithm is very efficient for large random 
graph instances. The relative sizes of the constructed FVS solutions by the 
local search algorithm are very close to
the predicted values of the replica-symmetric mean field theory, and they are
also very similar to the results obtained by the BPD message-passing algorithm.

It should be emphasized that the microscopic dynamical rules of our local 
search algorithm do not obey the detailed balance condition. 
These dynamical rules therefore
need to be appropriately modified if one is interested in the 
equilibrial FVS solutions at a given fixed temperature $T$.
We are currently using a modified set of microscopic dynamical rules to
study the spin glass phase transition of the
model proposed in \cite{Zhou-2013}. Such an equilibrium study 
will offer more physical insights on the cooling-rate dependent behaviors 
of Fig.~\ref{fig:evolution}.

\section*{Acknowledgement}
The numerical simulations were performed at the HPC computer cluster of
the authors' institute.
This work was partially supported by the National Basic Research Program
of China (No. 2013CB932804), the Knowledge Innovation Program of
Chinese Academy of Sciences (No.~KJCX2-EW-J02),
and the National Science Foundation of China
(grant Nos.~11121403, 11225526).
HJZ conceived research; SMQ performed research; HJZ wrote the paper.

\begin{appendix}
  
  \section{Proof of Theorem 1}
  Consider a generic legal list $L$ formed by some vertices of the graph $G$, 
  see Eq.~(\ref{eq:gList}). Let us consider the subgraph $F$ induced by all the
  vertices of list $L$ and all the edges between these vertices. Since
  $L$ is a legal list, every vertex $i\in L$ must satisfy the rank
  condition (\ref{eq:rc}). Consequently, every vertex in the subgraph $F$
  has at most one nearest neighbor with rank lower than that of
  itself. 

  This subgraph $F$ must be cycle-free. 
  We prove this statement by contradiction. Assume there is a cycle in
  the subgraph $F$ involving $k$ vertices:
  \begin{equation}
    \label{eq:ecycle}
    (i_1, i_2), (i_2, i_3), \ldots, (i_{k-1}, i_k), (i_k, i_1) \; .
  \end{equation}
  If the rank of vertex $i_1$ is lower than that of $i_2$, then due to the
  fact of vertex $i_2$ having at most one nearest neighbor with lower
  rank than itself, the rank of $i_{3}$ must be higher than that of $i_2$.
  Continuing this analysis along the cycle,  we obtain that the rank of
  vertex $i_k$ must be higher than that of $i_{k-1}$ but lower than that of
  $i_1$. But this is impossible since the rank of $i_{k-1}$ must be higher than
  that of $i_1$.
  Similarly, if the rank of vertex $i_1$ is higher than that of $i_2$, we will
  arrive at the contradicting results that the rank of $i_2$ is higher than
  that of $i_3$ while the rank of $i_3$ is higher than that of $i_1$.
  Because of these contradictions, the assumption of $F$ containing a cycle
  must be false. 

  Therefore $F$ must be a tree or a forest.
  Then the set $\Gamma$ formed by the vertices
  not contained in $L$ must be a feedback vertex set.

\section{Proof of Theorem 2}

Suppose the set $\Gamma$ is a FVS of a graph $G$. Then the subgraph
$F$ induced by all the vertices not included in $\Gamma$ must be cycle-free. 

If $F$ has only one connected component (i.e., being a tree), we can pick a
vertex (say $i$) of $F$ uniformly at random and specify this vertex as the
root of the tree subgraph $F$. We can then construct an ordered list $L$ using
all the vertices of $F$ in such a way: first the root vertex $i$, 
then all the vertices 
of unit path length to $i$ (in random order), followed by all the
vertices of path length two to $i$ (again in random order), ..., followed
by the remaining vertices of the longest path length to $i$ (in random order).
Obviously $L$ is a legal list with the ranking condition (\ref{eq:rc}) satisfied
for all its vertices.

If $F$ is a forest with two or more tree components, we can perform the
above-mentioned process for each of its tree components and then contatenate
the constructed ordered lists in a random order to form a whole ordered list $L$.
This list $L$ must also be a legal list.

\end{appendix}


\begin{thebibliography}{10}

\bibitem{Garey-Johnson-1979}
Garey M, Johnson DS (1979) {\em Computers and Intractability: A Guide to the
  Theory of NP-Completeness}.
\newblock (Freeman, San Francisco).

\bibitem{Festa-Pardalos-Resende-1999}
Festa P, Pardalos PM, Resende MGC (1999) in {\em Handbook of combinatorial
  optimization}, eds.{} Du DZ, Pardalos PM.
\newblock (Springer, Berlin, Germany), pp. 209--258.

\bibitem{Karp-1972}
Karp RM (1972) {\em Reducibility among combinatorial problems} eds.{} Miller E,
  Thatcher JW, Bohlinger JD.
\newblock (Plenum Press, New York), pp. 85--103.

\bibitem{Cook-1971}
Cook SA (1971) {\em The complexity of theorem-proving procedures} eds.{} Lewis
  PM et~al.
\newblock (ACM, New York), pp. 151--158.

\bibitem{Cao-Chen-Liu-2010}
Cao Y, Chen J, Liu Y (2010) On feedback vertex set new measure and new
  structures.
\newblock {\em Lect. Notes Comput. Sci.} 6139:93--104.

\bibitem{Kociumaka-Pilipczuk-2013}
Kociumaka T, Pilipczuk M (2013) Faster deterministic feedback vertex set
  (arXiv:1306.3566v1).

\bibitem{Guo-etal-2006}
Guo J, Gramm J, H\"{u}ffner F, Niedermeier R, Wernicke S (2006)
  Compression-based fixed-parameter algorithms for feedback vertex set and edge
  bipartization.
\newblock {\em J. Comput. System Sci.} 72:1386--1396.

\bibitem{Bafna-Berman-Fujito-1999}
Bafna V, Berman P, Fujito T (1999) A $2$-approximation algorithm for the
  undirected feedback vertex set problem.
\newblock {\em SIAM J. Discrete Math.} 12:289--297.

\bibitem{Razgon-2006}
Razgon I (2006) Exact computation of maximum induced forest.
\newblock {\em Lect. Notes Comput. Sci.} 4059:160--171.

\bibitem{Fomin-Gaspers-Pyatkin-2006}
Fomin FV, Gaspers S, Pyatkin AV (2006) Finding a minimum feedback vertex set in
  time $o(1.7548^n)$.
\newblock {\em Lect. Notes Comput. Sci.} 4169:184--191.

\bibitem{Bau-Wormald-Zhou-2002}
Bau S, Wormald NC, Zhou S (2002) Decycling numbers of random regular graphs.
\newblock {\em Random Struct. Alg.} 21:397--413.

\bibitem{Wang-Wang-Chang-2004}
Wang FH, Wang YL, Chang JM (2004) Feedback vertex sets in star graphs.
\newblock {\em Information Processing Letters} 89:203--208.

\bibitem{Jiang-Liu-Xu-2011}
Jiang W, Liu T, Xu K (2011) Tractable feedback vertex sets in restricted
  bipartite graphs.
\newblock {\em Lect. Notes Comput. Sci.} 6831:424--434.

\bibitem{Bayati-etal-2008}
Bayati M et~al. (2008) Statistical mechanics of steiner trees.
\newblock {\em Phys. Rev. Lett.} 101:037208.

\bibitem{Yeung-Saad-Wong-2013}
Yeung CH, Saad D, Wong KYM (2013) From the physics of interacting polymers to
  optimizing routes on the london underground.
\newblock {\em Proc. Natl. Acad. Sci. USA} 110:13717--13722.

\bibitem{Yeung-Saad-2013b}
Yeung CH, Saad D (2013) Networking--a statistical physics perspective.
\newblock {\em J. Phys. A: Math. Theor.} 46:103001.

\bibitem{Zhou-2013}
Zhou HJ (2013) Spin glass approach to the feedback vertex set problem.
\newblock {\em Eur. Phys. J. B} 86:455.

\bibitem{Kirkpatrick-etal-1983}
Kirkpatrick S, {Gelatt {Jr.}} CD, Vecchi MP (1983) Optimization by simulated
  annealing.
\newblock {\em Science} 220:671--680.

\bibitem{Galinier-Lemamou-Bouzidi-2013}
Galinier P, Lemamou E, Bouzidi MW (2013) Applying local search to the feedback
  vertex set problem.
\newblock {\em J. Heuristics} 19:797--818.

\bibitem{Biazzo-Braunstein-Zecchina-2012}
Biazzo I, Braunstein A, Zecchina R (2012) Performance of a cavity-method-based
  algorithm for the prize-collecting steiner tree problem on graphs.
\newblock {\em Phys. Rev. E} 86:026706.

\bibitem{He-Liu-Wang-2009}
He DR, Liu ZH, Wang BH (2009) {\em Complex Systems and Complex Networks}.
\newblock (Higher Education Press, Beijing).

\bibitem{Zhou-Lipowsky-2007}
Zhou HJ, Lipowsky R (2007) Activity patterns on random scale-free networks:
  global dynamics arising from local majority rules.
\newblock {\em J. Stat. Mech.: Theor. Exp.} p. P01009.

\bibitem{Zhou-Lipowsky-2005}
Zhou HJ, Lipowsky R (2005) Dynamic pattern evolution on scale-free networks.
\newblock {\em Proc. Natl. Acad. Sci. USA} 102:10052--10057.

\end{thebibliography}

\end{document}